\documentclass[twoside,11pt]{article}

\usepackage[abbrvbib, preprint]{jmlr2e}

\usepackage{booktabs, graphicx, subcaption}
\usepackage[inline]{enumitem}

\jmlrheading{1}{2022}{1-48}{11/21}{__/__}{____}{Divya Bhargavi, Karan Sindwani and Sia Gholami}

\ShortHeadings{Framework for 2D Ad placements in LinearTV}{Bhargavi, Sindwani and Gholami}
\firstpageno{1}

\begin{document}

\title{Framework for 2D Ad placements in LinearTV}

\author{\name Divya Bhargavi \email dbharga@amazon.com \\
  \name Karan Sindwani \email ksindwan@amazon.com \\
  \name Sia Gholami \email gholami@amazon.com \\
  \addr Amazon Web Services, CA USA}

\editor{__}

\maketitle

\begin{abstract}
  Virtual Product placement(VPP) is the advertising technique of digitally placing a branded object into the scene of a movie or TV show. This type of advertising provides the ability for brands to reach consumers without interrupting the viewing experience with a commercial break, as the products are seen in the background or as props. Despite this being a billion-dollar industry, ad rendering technique is currently executed at post production stage, manually either with the help of VFx artists or through semi-automated solutions. In this paper, we demonstrate a fully automated framework to digitally place 2-D ads in linear TV cooking shows captured using single-view camera with small camera movements. Without access to full video or production camera configuration, this framework performs the following tasks
  \begin{enumerate*}[label=(\roman*)]
    \item identifying empty space for 2-D ad placement
    \item kitchen scene understanding
    \item occlusion handling
    \item ambient lighting and
    \item ad tracking.
  \end{enumerate*}
\end{abstract}








\section{Introduction}\label{sec:introduction}
Rendering a realistic 3-D ad object requires knowledge of 3-D scene through camera calibration process or devices that can record camera parameters~\citep{4,5} or, depth and scale of objects in the scene, light sources and their location in 3-D as well as the knowledge of foreground/background segmentation maps. The reason to explore 2-D ad rendering as opposed to 3-D is as follows:
\begin{enumerate}
    \item Streaming platforms purchase videos from 3rd party vendors who most often don’t have access to camera parameters used for video production, for 3-D scene understanding.
    \item There is no object/reference structure of known dimensions to calibrate scale for 2-D to 3-D point transformations and
    \item Live/real-time ad rendering applications with single view camera precludes one from using long-form videos to use techniques like Structure from Motion (SfM) and multi-view stereo, as we process the frames sequentially~\citep{6, 7, 8, 9}.
\end{enumerate}
Existing computer vision based VPP approaches are either semi-automatic (requiring user input for ad location, occlusion handling, adjust ad rendering)~\citep{10} or automatic  with ad replacement on specific targets like billboards~\citep{12}. With the lack of standardized data-sets and opens-source repositories, the task of quickly prototyping an end-to-end solution for a potential commercial use is harder.

Our contributions are:
\begin{enumerate}
    \item We develop an end-to-end solution that automatically inserts 2-D ads into cooking show videos.
    \item We introduce 3 different ways of detecting empty spaces on indoor scene walls.
    \item We explore line-segmentation based models for perspective alignment.
    \item We build a framework that could generalize to 2-D ad insertions in any type of scene with minimal camera movement.
\end{enumerate}

\section{Related Work}\label{sec:related-work}

\subsection{Inverse Rendering in Indoor Scenes}\label{subsec:inverse-rendering-in-indoor-scenes}
Inverse rendering in indoor scenes is the task of decomposing a single RGB scene into material (albedo and roughness), geometry (depth and normal), and spatially-varying lighting of the scene with applications in object placement and editing scene material and lighting. While the literature in this domain~\citep{11,18,19} addresses most product placement challenges on 3D scene understanding and lighting, there is a dearth of open-source implementations for commercial use as well as documentation on how these generalize when tested on different scenes within long form video for consistent estimates. Additionally, for an automated pipeline, problems like identifying empty space and tracking ad location should still be addressed.

\subsection{Plane Detection}\label{subsec:plane-detection}
Plane detection is the task of identifying planar structures in scenes. With the ubiquitous use of Convolution Neural Networks (CNNs) in computer vision task, there has been promising increase in literature in considering this task as a segmentation task. Models like PlaneNet, PlaneRecover~\citep{13, 15,16} attempt to segment a fixed number of planes in an image but fail to generalize on different scenes and smaller plane structures. PlaneRCNN attempts to improve on the issues raised previously by detecting planar regions and reconstructing a piecewise planar depth-map from a single RGB image. This however requires camera intrinsic parameters for refinement and 3D reconstruction. In our work, we use plane detection models to identify and delineate different wall structures in the background for empty space identification.

\subsection{Instance Segmentation}\label{subsec:instance-segmentation}
Instance segmentation is the task of detecting and disambiguating distinct objects in an image. Models in this domain exist in two paradigms namely one-stage and two-stage. Two stage models first identify a set of object proposals and then identify segmentation maps by differentiating foreground-background~\citep{23,24,25} . One-stage methods~\citep{26,27}  could be anchor based or anchor that use related parallel design and dense prediction network to achieve comparable accuracy as two-stage models. For developing a prototype, we chose to work with two-stage models that have better accuracy compared to one-stage models. Since we were prioritizing an accurate pipeline for our prototype, we chose a two-stage Mask-RCNN based backbone for our experiments.

\subsection{Light Estimation}\label{subsec:light-estimation}
Light Estimation is a sub-task in inverse rendering domain that learns to disambiguate between properties of light, materials and their interaction in 3D space(reflectance, geometry and shape). While outdoor lighting setting is simplified with no assumptions on spatial variations in light~\citep{29, 30}, it becomes an important problem to solve for indoor settings. Methods to solve lighting for complex indoor scenes have evolved from learning light environment maps, parametric models, spherical lobe designs to consistent 3D spatially varying HDR (high dynamic range) light estimation~\citep{21,22,28}. In the absence of open source pre-trained models, camera intrinsic and stereo images, none of the deep learning methods apply for our use case. We use classical CV methods that learn global illumination properties and applies them on to an ad image.

\subsection{Key-point Detection and Description}\label{subsec:keypoint-detection-and-description}
Key-point Detection and Description is the task of detecting stable interest points in an image and encoding them as descriptors that contain their properties. This is one of the fundamental tasks in SfM, simultaneous localization and mapping (SLAM), image matching and vision localization among others. It has evolved from classical algorithms like SIFT, ORB~\citep{31,32} that were hugely successful to local detector based methods~\citep{14}, to Transformer based models that capture global features through attention mechanism~\citep{33}. We explore algorithms across all the variations along with different key-point matching and homography estimation/outlier detection algorithms~\citep{36,37}. by comparing them using re-projection error.


\section{Approach}\label{sec:approach}
\begin{figure}[ht]
    \centering
    \includegraphics[width=0.8\linewidth]{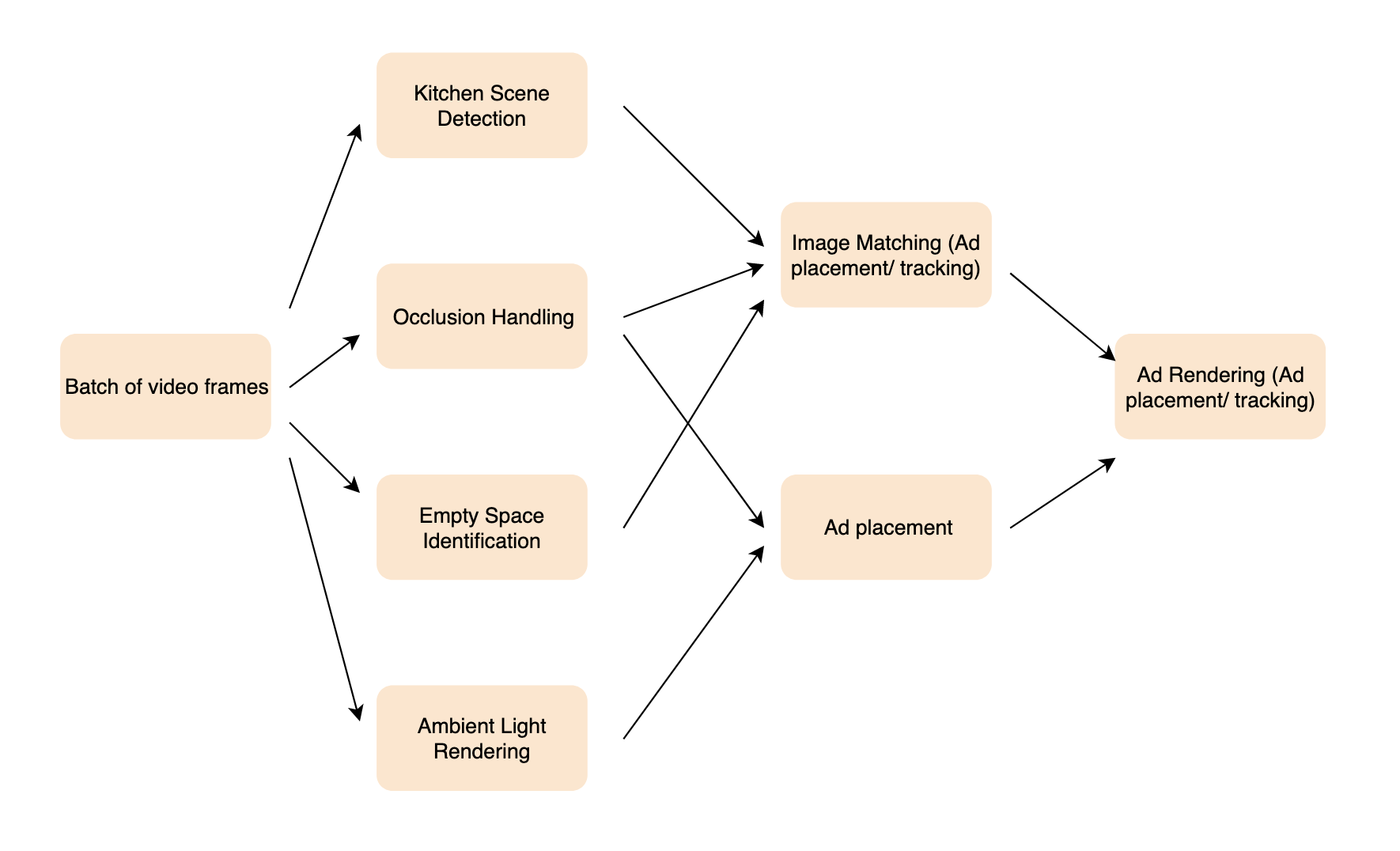}
    \caption{Automated Product Placement pipeline that processes a batch of frames}
    \label{fig:pipeline}
\end{figure}

Our solution consists of 6 key steps as show in Figure~\ref{fig:pipeline}.  The following sections cover each of these steps in detail.

\subsection{Identifying suitable placement location}\label{subsec:identifying-suitable-placement-location}
The objective is to develop a Machine Learning (ML) model that can identify suitable placement locations for 2-D objects (posters, Ad images) on a wall. Suitable placement locations can be on other kitchen objects as well (Microwave, oven, refrigerator) but these were not considered in the scope of this work. Models used for this task were selected on the basis of them being state-of-the-art for a given task or doesn't require camera parameters. We experimented 2 different strategies: One was a rule-based approach using pre-trained models while the other involved training a custom model on the data.

\subsubsection{Rule-based approach}\label{subsubsec:rule-based-approach}
\begin{figure}[ht]
    \centering
    \begin{minipage}{.5\textwidth}
      \centering
      \includegraphics[width=0.9\linewidth]{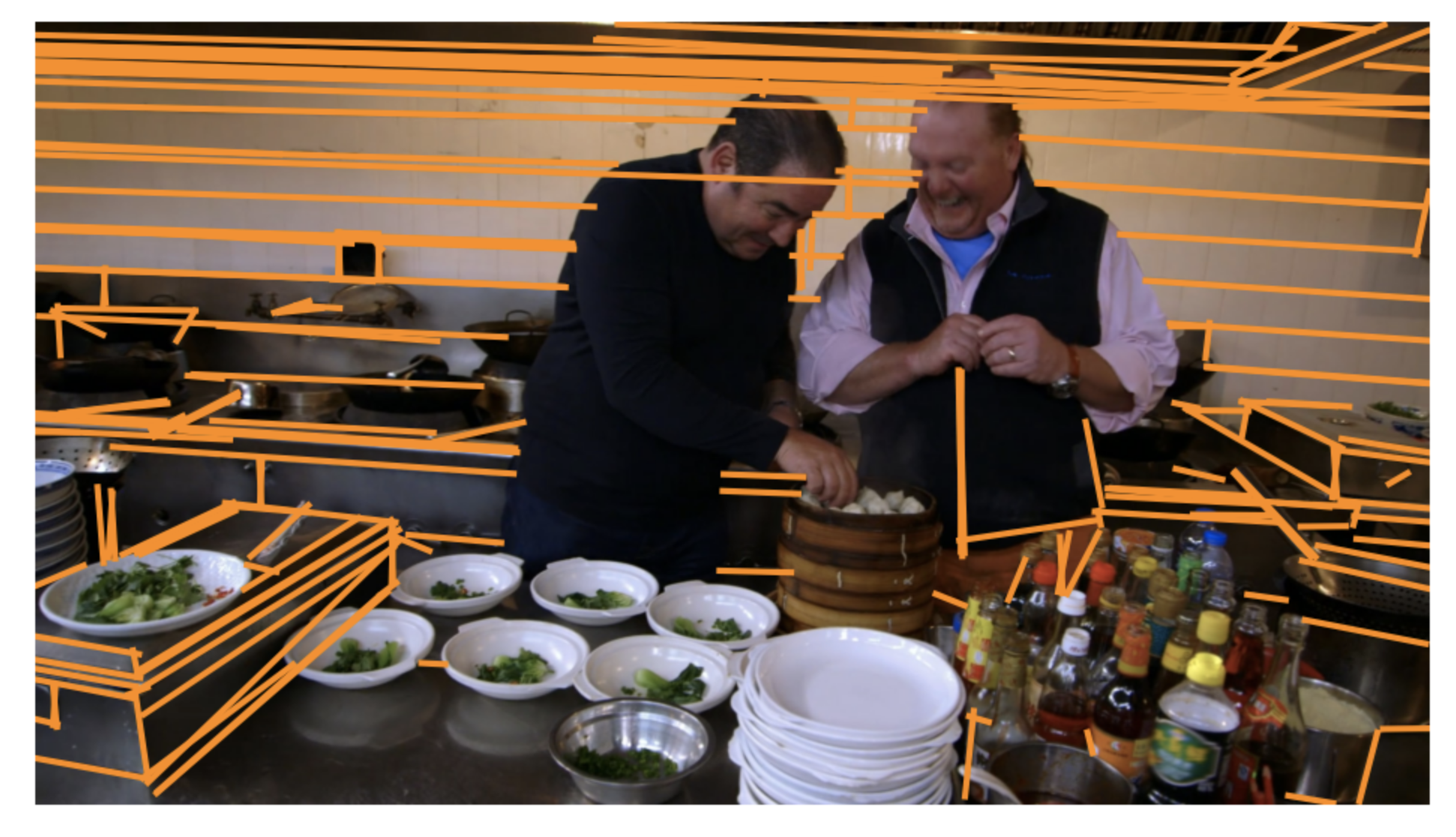}
      \caption{LETR output}
      \label{fig:letr-output}
    \end{minipage}%
    \begin{minipage}{.5\textwidth}
      \centering
      \includegraphics[width=0.9\linewidth]{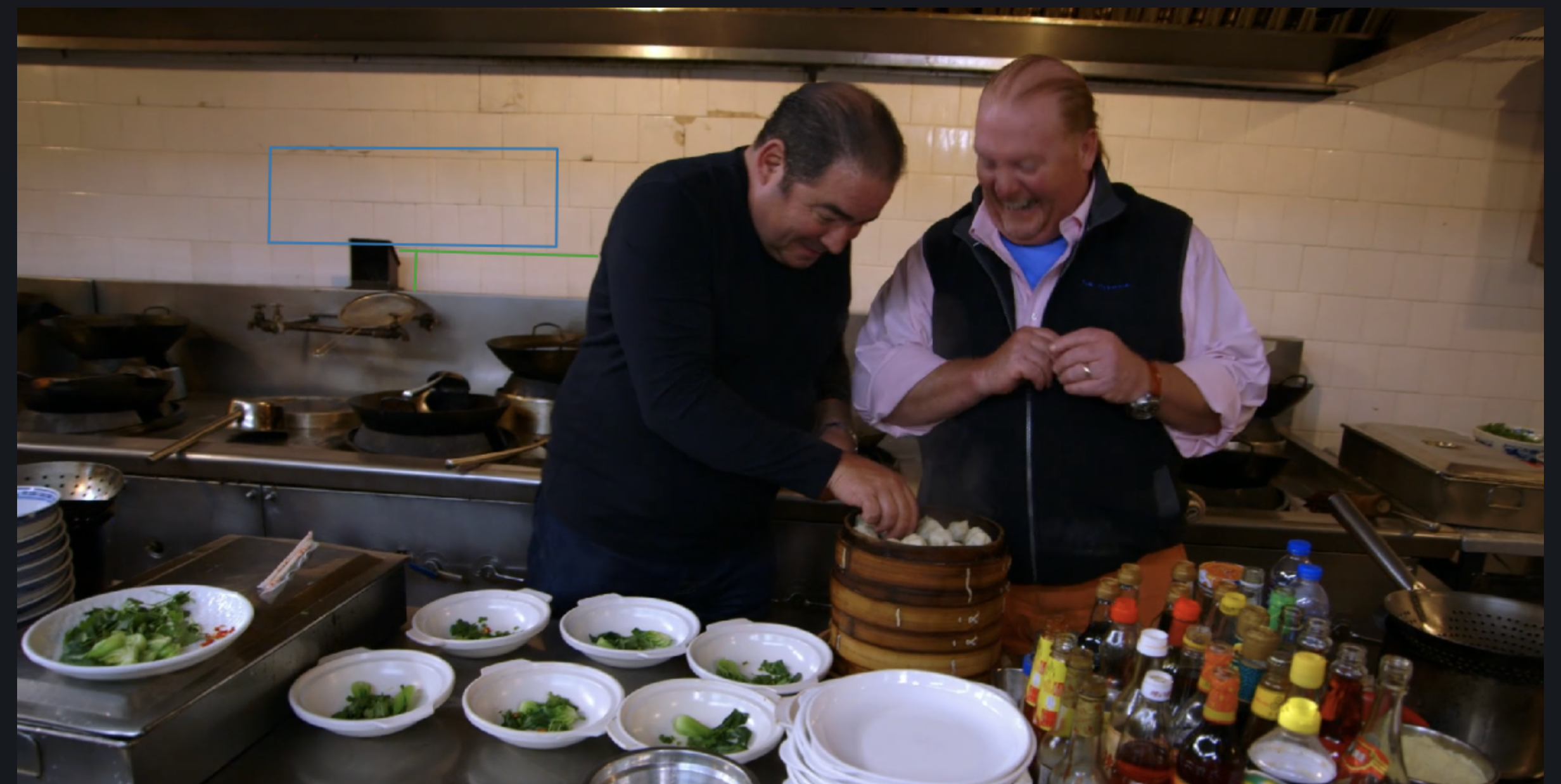}
      \caption{Lines closer to bounding box proposal}
      \label{fig:closer-lines}
    \end{minipage}
    \end{figure}

The rule-based approach is executed sequentially in the following order:
\begin{enumerate}

    \item First detecting wall using pre-trained models in Detectron2~\citep{38} library (see figure~\ref{fig:wall-detection}). We use a panoptic-FPN segmentation~\citep{40} model pre-trained on ADE dataset~\citep{39} and filter on the wall classes. We also detect distinct planar surfaces in the video frame using PlanarReconstruction~\citep{17} model to disambiguate different folds of the wall (see figure~\ref{fig:plane-detection}).
    
    \begin{figure}[ht]
    \centering
    \begin{minipage}{.5\textwidth}
      \centering
      \includegraphics[width=0.9\linewidth]{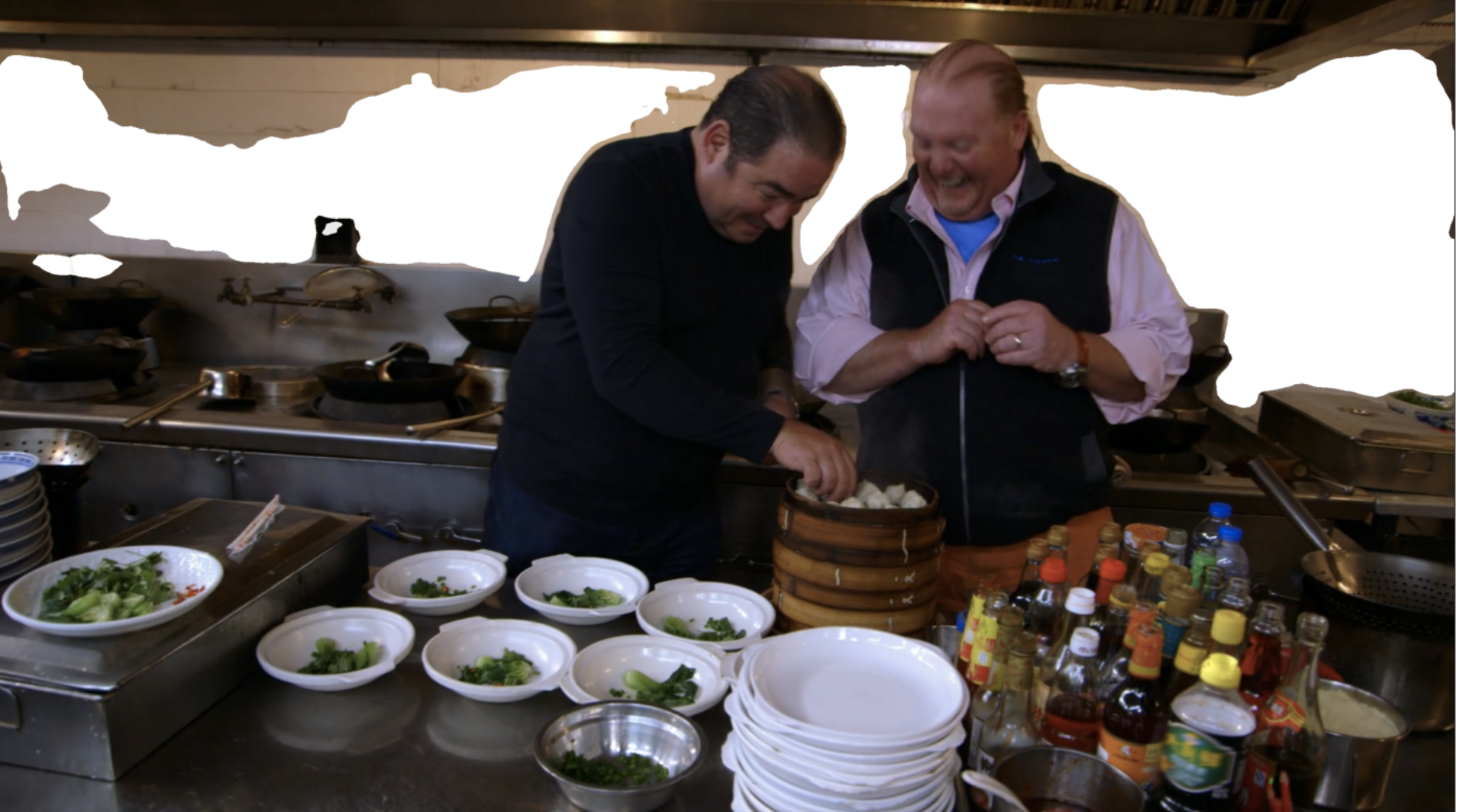}
      \captionof{figure}{Wall detection}
      \label{fig:wall-detection}
    \end{minipage}%
    \begin{minipage}{.5\textwidth}
      \centering
      \includegraphics[width=0.9\linewidth]{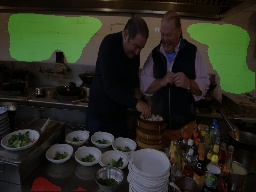}
      \captionof{figure}{Plane detection}
      \label{fig:plane-detection}
    \end{minipage}
    \end{figure}

    \item We then generate an empty space mask using the intersection of the results from wall segmentation and plane detection models (see figure~\ref{fig:empty-space-mask}).The segmentation models/mask don’t have the information to distinguish different blobs in the mask.

    \item We use a region proposal function from scipy~\citep{41} package to get region proposals/bounding box for each blob(see Figure~\ref{fig:region-proposals}). The above-mentioned parts of the rule-based pipeline return suitable placement locations/regions in an image but these locations/regions may not be prospectively aligned.

    \begin{figure}[ht]
    \centering
    \begin{minipage}{.5\textwidth}
      \centering
      \includegraphics[width=0.9\linewidth]{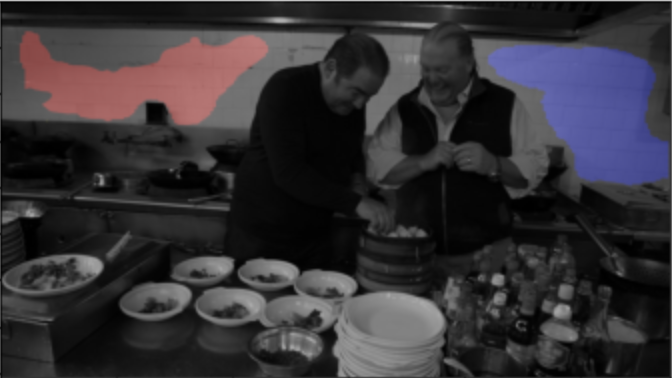}
      \captionof{figure}{Empty Space mask}
      \label{fig:empty-space-mask}
    \end{minipage}%
    \begin{minipage}{.5\textwidth}
      \centering
      \includegraphics[width=0.9\linewidth]{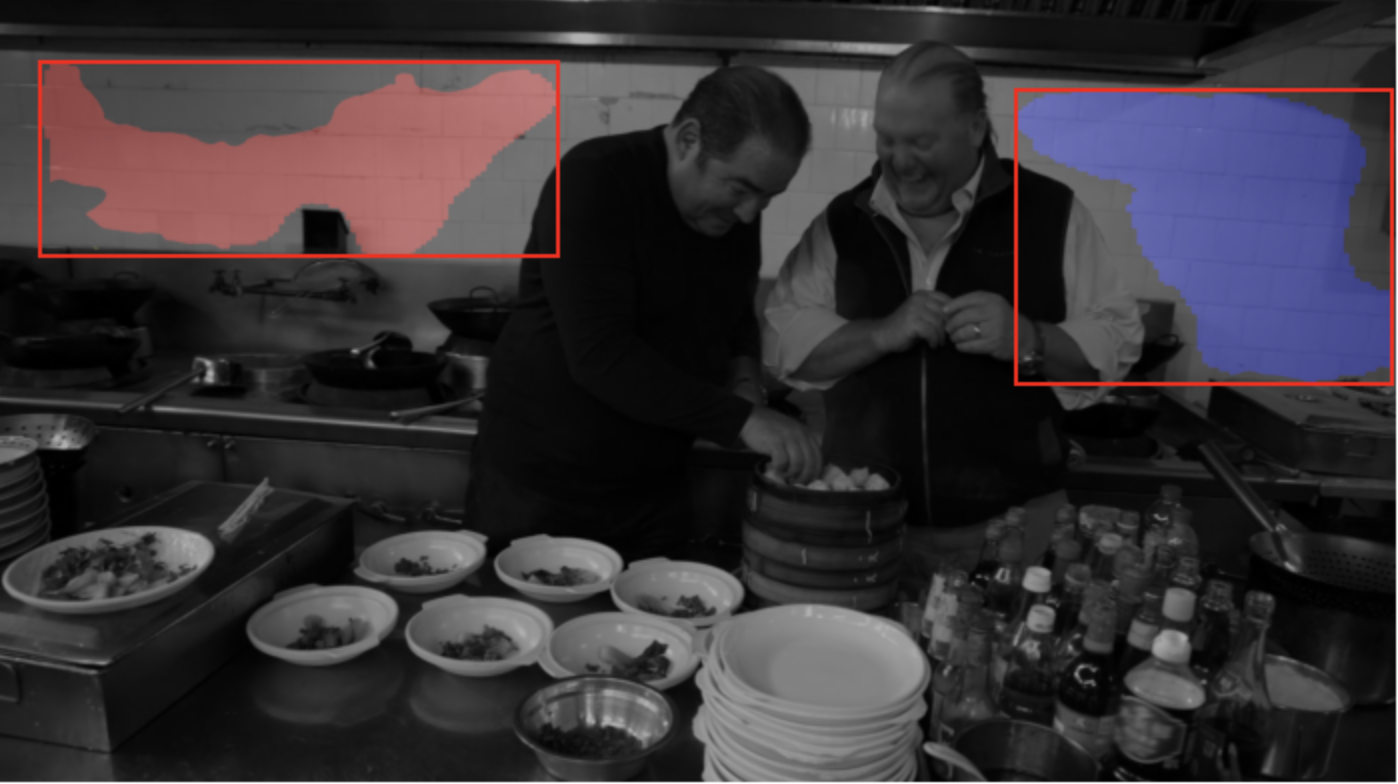}
      \captionof{figure}{Region Proposals}
      \label{fig:region-proposals}
    \end{minipage}
    \end{figure}

    \item We align the bounding boxes by:
    \begin{enumerate}
        \item We use LETR(Line Segment Detection Using Transformers without Edges) model~\citep{42} to generate lines.(see figure~\ref{fig:letr-output}).
        \item Then we classify these lines as vertical or horizontal by measuring slope of the line.
        \item The next step in order to align the region to wall line segments is to find the closest vertical and horizontal lines. There are multiple ways of computing a distance between a region and a line segment. We took an approach which calculates the distance between the center of the region and the endpoints on the line segment and took the pair with the minimum distance. (see figure~\ref{fig:closer-lines})
        \item Compute adjusted region points with slope of LETR line segments. Given that a point ($x_1$, $y_1$) is at distance $d$ away from ($x$, $y$). We can generate $x_1$ and $y_1$ co-ordinates using the following formulae.
        \begin{equation}
            \label{eq:2}
            r = \sqrt{1+m^2}
        \end{equation}
        \begin{equation}
            \label{eq:3}
            (x_1,y_1) = (x + \frac{d}{r}, y + \frac{d.m}{r})
        \end{equation}
    \end{enumerate}

\end{enumerate}

\subsubsection{Custom model approach}\label{subsubsec:custom-model-approach}
The rule-based approach utilizes 3 different models which could potentially lead to latency and cascading error issues. Hence we also tested 2 different custom modeling approaches. The \textbf {Polygon Regression} method directly regressed to predict a perspective aligned bounding box using Yolov5~\citep{43} model. The  \textbf {Instance Segmentation approach} ,  identifies patches/segment on wall for ad placement. For this approach, a Mask-RCNN~\citep{23} was trained. We compare both these approaches based on the box Intersection over Union (IoU) and angle of deviation with the ground truth bounding box (bbox) lines.


\subsection{Kitchen Scene Identification}\label{subsec:kitchen-scene-identification}
Kitchen scene detection is a sub-task of “Identifying empty space”. In addition to identifying just an empty space in an image, the VPP pipeline should also be able to discern if the frame being captured is within a kitchen (project objective) as opposed to outdoors or other areas, and render image accordingly.  We use pre-trained CV models with rule-based approach to classify whether a scene is shot from kitchen or elsewhere. We define a scene as “kitchen scene” when a person is clearly visible (confidence scores above 0.95) and the surrounding area has kitchen related artifacts (used relevant shortlisted classes). We tested 3 different pre-trained models: Amazon Rekognition (https://aws.amazon.com/rekognition/), Faster R-CNN~\citep{44} and RetinaNet~\citep{45}. We chose the models by considering SOTA accuracy on person and kitchen related item classification metrics.

\subsection{Occlusion Handling}\label{subsec:occlusion-handling}
In the absence of foreground-background maps and camera parameters, we formulated the 2-D VPP ad object to be on walls which is mostly on the background and it is reasonable to say any object that occludes its view will be on foreground. We only test the occlusion by humans as it is impossible to produce segmentation masks for unknown objects that the person in cooking shows might interact with. We benchmark semantic segmentation, instance segmentation and panoptic segmentation models against Human Segmentation Data~\citep{46} that had high definition masks of humans with different posture and background on IoU scores.

\subsection{Ambient Light Rendering}\label{subsec:ambient-light-rendering}
The goal of this task is to match the light (perception of lighting) of an advertisement image to a background image. Since there are no publicly available datasets or models and we did not have any labeled data, we did not use machine learning and leveraged classical CV approaches for this task. We experimented with the following methods:

\subsubsection{Image Brightness Matching}\label{subsubsec:image-brightness-matching}
In this method, we try to match the brightness of the advertisement image to the background image. This method is based on brightness calculation as presented in Szeliski, 2010~\citep{47}. First, we calculate the brightness of the background image and then adjust the brightness of the ad to match that value.

\begin{equation}
    g(x) = \alpha * f(x) + \beta
\end{equation}
$\alpha$ and $\beta$ are contrast and brightness respectively

\subsubsection{Color Transfer}\label{subsubsec:color-transfer}
This method is based on the work presented in Color transfer between images paper published by Reinhard et al~\citep{48}. This method uses statistical analysis to impose one image’s color characteristics on another using Lab color space and the mean and standard deviation of each L, a, and b channel, respectively.

\subsubsection{LAB Light Transfer}\label{subsubsec:lab-light-transfer}
In this method, we attempt to transfer the background image’s light (in LAB format) to the advertisement image based on the work presented in~\citep{48} by Reinhard et al. The primary difference between this method and ‘Color Transfer’ method presented above is that this method a and b channels do not change and only L channel will be transferred.

\subsubsection{Histogram Matching}\label{subsubsec:histogram-matching}
In this method, we attempt to match the ad's image’s histogram to the background image. This method is based on the works of Gonzalez et al~\citep{49} and is a generalized version of well-known histogram equalization method. The algorithm starts by finding a set of unique pixel values and their corresponding indices and counts. Then it takes the cumulative sum of the counts and normalizes by the number of pixels to get the empirical cumulative distribution function for the background and ad image.

\subsection{Ad placement}\label{subsec:ad-placement2}
The goal of this task is to place the ad image in the video, given its location coordinates and segmentation maps of occluding objects. We developed a computer vision module that places an Ad image on to the video frame. In addition to the binary map of human segmentation and ambiently lit ad image, we used OpenCV~\citep{50} based \emph{getPerspectiveTransform} function to learn the transform from ad image to the placement location on the image. We tested this method as opposed to simply pasting the image (for rectangular empty space locations only) on the empty location to handle “perspective” adjusted quadrilaterals of any shape in the future. The learnt transformation matrix will warp the Ad image according to the empty space location dimensions. Before rendering the image, we mask out those regions of the ad that are occluded by humans in the scene using segmentation maps.

\subsection{Ad tracking}\label{subsec:ad-tracking}
The objective for this task is to track the ad in a video for consistent and realistic rendering in the same location. We developed a computer vision module that tracks the location of ad in consecutive frames given previous video frame and its location coordinates.  This module uses keypoint detector/descriptor, keypoint feature matcher and homography estimation functions. We have to note that this work was done on the premise that the camera parameters are unknown to learn 3D world to 2-D video frame mapping.

Tracking the location of Ad in consequent frames consists of the following tasks:
\begin{enumerate}
    \item Mask out occluding humans in image (refer to~\ref{subsec:occlusion-handling} task from above)
    \item Detection and Description: This involves understanding key features in an image and generating a feature-vector/ embedding. The models tested were the following:
    \begin{enumerate}
        \item Classical CV (OpenCV) : ORB, SIFT~\citep{31,32}
        \item Deep Learning: SuperPoint (implementation)~\citep{14}, Kornia library~\citep{33}.
    \end{enumerate}
    \item Remove features in and around occluding human. This is done so that the tracking is based background objects than the human features.
    \item Feature Matching: This involves matching the features generated in both the images for correspondence. We tested Brute Force, Single Nearest Neighbor, Mutual Nearest Neighbor, FGINN (1st geometrically inconsistent nearest neighbor ratio)~\citep{51} and GMS(Grid-based Motion)~\citep{52}.
    \item Outlier Detection: This involves removing the outliers in feature matching using thresholds on “matching” metric. We tested RANSAC and MAGSAC~\citep{36,37}.
    \item Learn the homography matrix through OpenCV functions: This involves learning the transformation matrix (approximation of mild camera movement) between previous and current image.
    \item Get location coordinates: This involves applying the transformation on previous image Ad location coordinates to get new location coordinates for current image. We benchmark these algorithms using re-projection error.
\end{enumerate}

\section{Experimental Results}\label{experimental-results}

\subsection{Data-set}\label{subsec:dataset}
\begin{figure}[ht]
    \begin{center}
    \includegraphics[width=0.7\linewidth]{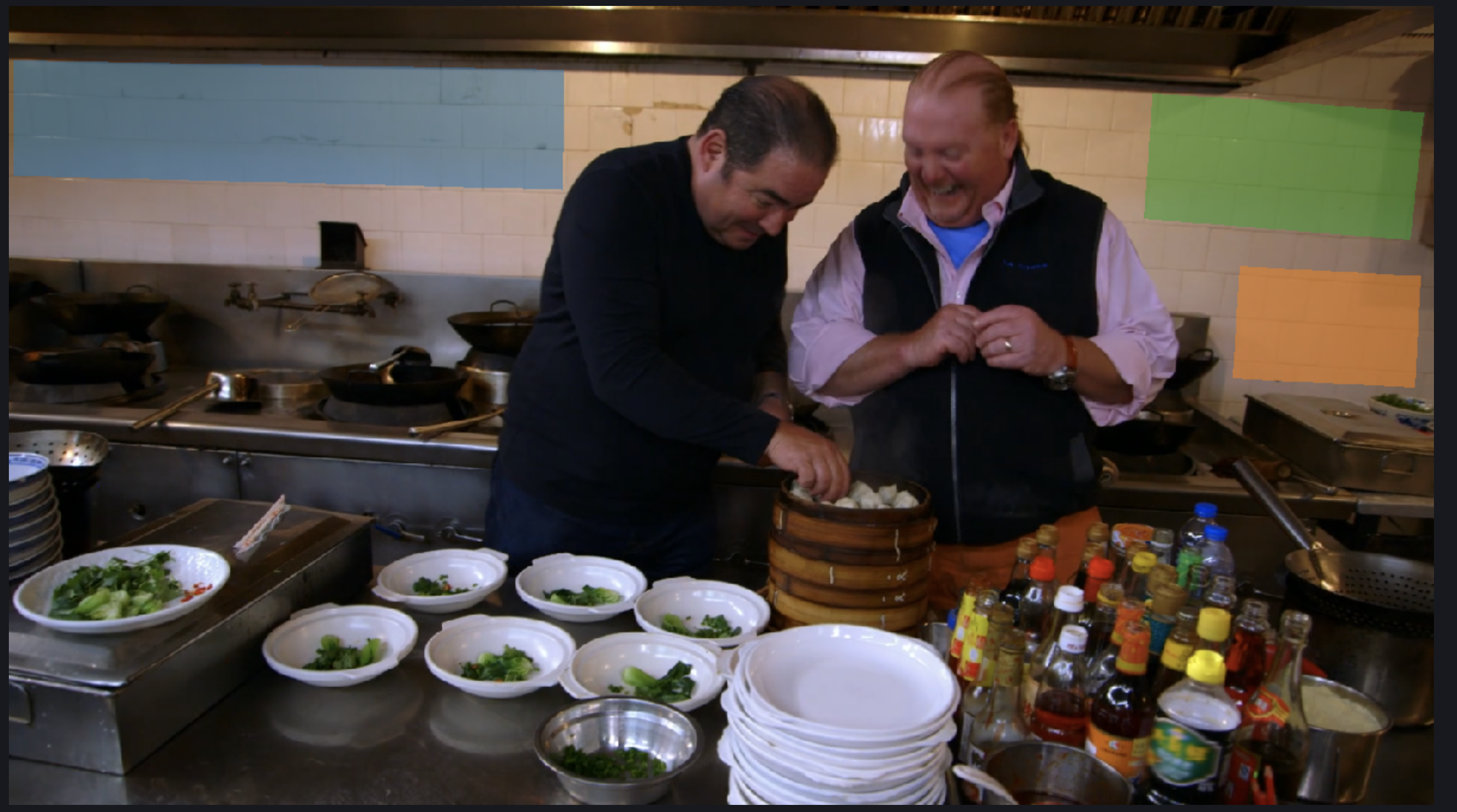}
    \caption{Ground Truth Label}
    \label{fig:ground-truth-label}
    \end{center}
\end{figure}

Our data-set consists of 25 cooking shows in mp4 format videos with resolution of 288x512. We sampled and labelled 1200 images (see figure~\ref{fig:ground-truth-label}). The video frames were labeled using the following mechanism: 
\begin{enumerate}
\item An image was labeled if it had a kitchen scene with empty space on walls and had the presence of a person in full view
\item An image was not labeled/discarded if it was a close up of a cooking scene, a non-kitchen scene or the scene did not have any empty space on the wall
\item An empty space was labeled using a polygon based bounding box
\item 2-3 large empty spaces were marked for each image. Additionally, we augmented the labelled images with Gaussian Noise,  Optical Distortion, Channel Shuffle and Random Cropping techniques.
\end{enumerate}

\subsection{Identifying suitable placement location}\label{subsec:identifying-suitable-placement-location2}

\subsubsection{Rule-based approach}\label{rule-based-approach-exp}
The rule based approach was evaluated based on qualitative results is discussed in~\ref{subsubsec:rule-based-approach}. We observed several challenges with rule-based approach. The results from wall detection model were inconsistent and consists of a significant amount of false positives. The results from PlanarReconstruction model which were used to disambiguate different folds wasn't accurate enough for our task.The results from the rule-based pipeline are sensitive to the slightest camera movement. Alignment pipeline is highly dependent on the wall background. It achieves higher performance on brick backgrounds and degrades on solid wall types.

\subsubsection{Custom model approach}
Table represents benchmarking of custom models for identifying suitable locations on our annotated dataset with respect to IoU (Intersection over Union) and Angle deviation between all 4 quadrilateral lines of ground truth and model predictions. Yolo-v5 (Polygon regression model) is relatively better at predicting empty spaces with low/ minimal overlap/occlusion with real life objects. However, the Mask-RCNN (custom segmentation) model gave a lot more candidate spaces with lower deviation in perspective compared to Yolo-v5 on our ground truth. After qualitative (section~\ref{subsubsec:custom-model-approach}) and quantitative analysis (table~\ref{tab:custom-model-results}) of both models, we used instance segmentation approach to build the automated VPP pipeline.
%

\begin{table}[ht]
  \caption{Custom model results}
  \label{tab:custom-model-results}
    \centering
    \small
    \begin{tabular}{lllll}
        \toprule
        Model & Approach & Avg IOU & Avg angle deviation  & GT box overlap \\
        \midrule
        Yolo-v5 & Polygon Regression & 0.56 & 3.27 & 40/42 \\
        Mask-RCNN & Instance Segmentation & 0.52 & 3 & 37/42 \\
        \bottomrule
    \end{tabular}
\end{table}

\subsection{Kitchen Scene Classification}\label{subsec:kitchen-scene-classification}
We define a positive classification of kitchen scene when a person is detected with a confidence of 90\% or above and the image contains kitchen artifacts like ‘bottle’, ‘wine glass’, ‘cup’, ‘fork’, ‘knife’, ‘spoon’ and ‘bowl’ with a confidence of 80\%. The 95\% threshold filters out scenarios when the camera focuses on cooking pan or close up of a region in the kitchen when the person could be partially visible or completely out of scene. 80\% threshold for kitchen artifacts was decided based on qualitative evaluation. We used the same dataset as empty space identification model. All the images where we marked an empty space box or marked a kitchen scene with no empty space tag were considered positive classes. Rest of the images were marked as negative class. RetinaNet had the highest accuracy. This model has a smaller architecture compared to Faster R-CNN making it a better candidate for latency related constraints.

\begin{table}[ht]
  \caption{Scene classification results}
  \label{tab:scene-classification-results}
    \centering
    \small
    \begin{tabular}{ll}
        \toprule
        \textbf{Model} & \textbf{Accuracy} \\
        \midrule
        Retina-Net & 0.926 \\
        Faster-RCNN & 0.852 \\
        Amazon Rekognition & 0.822 \\
        \bottomrule
    \end{tabular}
\end{table}

\subsection{Occlusion Handling}
We quantitatively compare latency benchmarks and IoU results~\citep{46} of pre-trained Image segmentation models in table~\ref{fig:table_8} and table~\ref{tab:inference-time-benchmark} . We observed the following key takeaways: Semantic Segmentation models have  better IoU performance than Panoptic and Instance segmentation models. Instance/Panoptic Segmentation models performed 2x better in GPU/CPU inference latency than segmentation models. Models trained COCO, VOC dataset perform better in human segmentation than model trained over ADE dataset. Based on our qualitative evaluation on the dataset in~\ref{fig:img_15}, we noticed Mask-RCNN model is unable to produce a prediction across all image resolutions and Panoptic segmentation models perform better than Mask-RCNN models across all image resolutions.

\begin{table}[ht]
  \caption{Inference Time benchmark - CPU and GPU Latency}
  \label{tab:inference-time-benchmark}
    \centering
    \small
    \begin{tabular}{lllll}
        \toprule
        \textbf{Method} & \textbf{Model} & \textbf{Image Size} & \textbf{CPU} & \textbf{GPU} \\
        \midrule
        Panoptic Segmentation & Panoptic fpn R50 & 2160 x 3840 & 7.497 & 0.178 \\
                                                  &  & 140 x 250 & 3.350 & 0.078 \\
                                                  &  & 281 x 500 & 3.521 & 0.080 \\
                                                  &  & 562 x 1000 & 3.598 & 0.085 \\
                                \cline{2-5}
                                & Panoptic fpn R5101 & 2160 x 3840 & 8.082 & 0.188 \\
                                                  &  & 140 x 250 & 4.148 & 0.090 \\
                                                  &  & 281 x 500 & 4.085 & 0.094 \\
                                                  &  & 562 x 1000 & 4.248 & 0.101 \\
        \midrule
        Instance Segmentation & Mask RCNN R50 & 2160 x 3840 & 4.831 & 0.165 \\
                                                  &  & 140 x 250 & 5.158 & 0.095 \\
                                                  &  & 281 x 500 & 4.977 & 0.097 \\
                                                  &  & 562 x 1000 & 4.985 & 0.103 \\
                                \cline{2-5}
                                & Mask RCNN R101 & 2160 x 3840 & 3.701 & 0.172 \\
                                                  &  & 140 x 250 & 3.620 & 0.082 \\
                                                  &  & 281 x 500 & 3.751 & 0.080 \\
                                                  &  & 562 x 1000 & 3.671 & 0.083 \\
                                \cline{2-5}
                                & Mask RCNN X101 & 2160 x 3840 & 6.206 & 0.202 \\
                                                  &  & 140 x 250 & 5.746 & 0.126 \\
                                                  &  & 281 x 500 & 5.859 & 0.128 \\
                                                  &  & 562 x 1000 & 5.744 & 0.131 \\
        \midrule
        FCN Semantic Segmentation & FCN ResNet101 & 2160 x 3840 & 94.25 & 2.610 \\
                                                  &  & 140 x 250 & 0.440 & 0.311 \\
                                                  &  & 281 x 500 & 1.250 & 0.458 \\
                                                  &  & 562 x 1000 & 5.928 & 0.424 \\
        \midrule
        PSP Semantic Segmentation & PSP ResNet101 & 2160 x 3840 & 94.800 & 2.063 \\
                                                  &  & 140 x 250 & 0.504 & 0.153 \\
                                                  &  & 281 x 500 & 1.414 & 0.080 \\
                                                  &  & 562 x 1000 & 6.367 & 0.155 \\
        \midrule
        DeepLab V3 Semantic Segmentation & DeepLab ResNet101 & 2160 x 3840 & 95.300 & 2.143 \\
                                                  &  & 140 x 250 & 0.447 & 0.095 \\
                                                  &  & 281 x 500 & 1.470 & 0.076 \\
                                                  &  & 562 x 1000 & 6.384 & 0.161 \\
         \bottomrule
    \end{tabular}
\end{table}

\begin{table}[ht]
  \caption{Comparison of models on Human Segmentation dataset}
  \label{fig:table_8}
    \centering
    \small
    \begin{tabular}{llll}
        \toprule
        \textbf{Dataset/Framework} & \textbf{Segmentation Type} & \textbf{Model Name} & \textbf{IoU} \\
        \midrule
        COCO/Detectron2 & Panoptic  & panoptic\_fpn\_R\_50\_3x & 0.907 \\
COCO/Detectron2 & Panoptic  & panoptic\_fpn\_R\_101\_3x & 0.908 \\
COCO/Detectron2 & Instance  & mask\_rcnn\_R\_101\_FPN\_3x & 0.908 \\
COCO/Detectron2 & Instance  & mask\_rcnn\_X\_101\_32x8d\_FPN\_3x & 0.907 \\
VOC/GluonCV & Semantic  & fcn\_resnet101 & 0.916 \\
VOC/GluonCV & Semantic  & psp\_resnet101 & 0.920 \\
VOC/GluonCV & Semantic  & deeplab\_resnet101 & 0.927 \\
COCO/GluonCV & Semantic  & fcn\_resnet101 & 0.924 \\
COCO/GluonCV & Semantic  & psp\_resnet101 & 0.927 \\
COCO/GluonCV & Semantic  & deeplab\_resnet101 & 0.928 \\
ADE/GluonCV & Semantic  & fcn\_resnet101 & 0.710 \\
ADE/GluonCV & Semantic  & psp\_resnet101 & 0.716 \\
ADE/GluonCV & Semantic  & deeplab\_resnet101 & 0.737 \\
    \bottomrule
    \end{tabular}
\end{table}

\subsection{Ambient Light Rendering}\label{subsec:ambient-light-rendering-exp}
\begin{figure}
\centering
\begin{subfigure}{0.4\textwidth}
    \includegraphics[width=\textwidth]{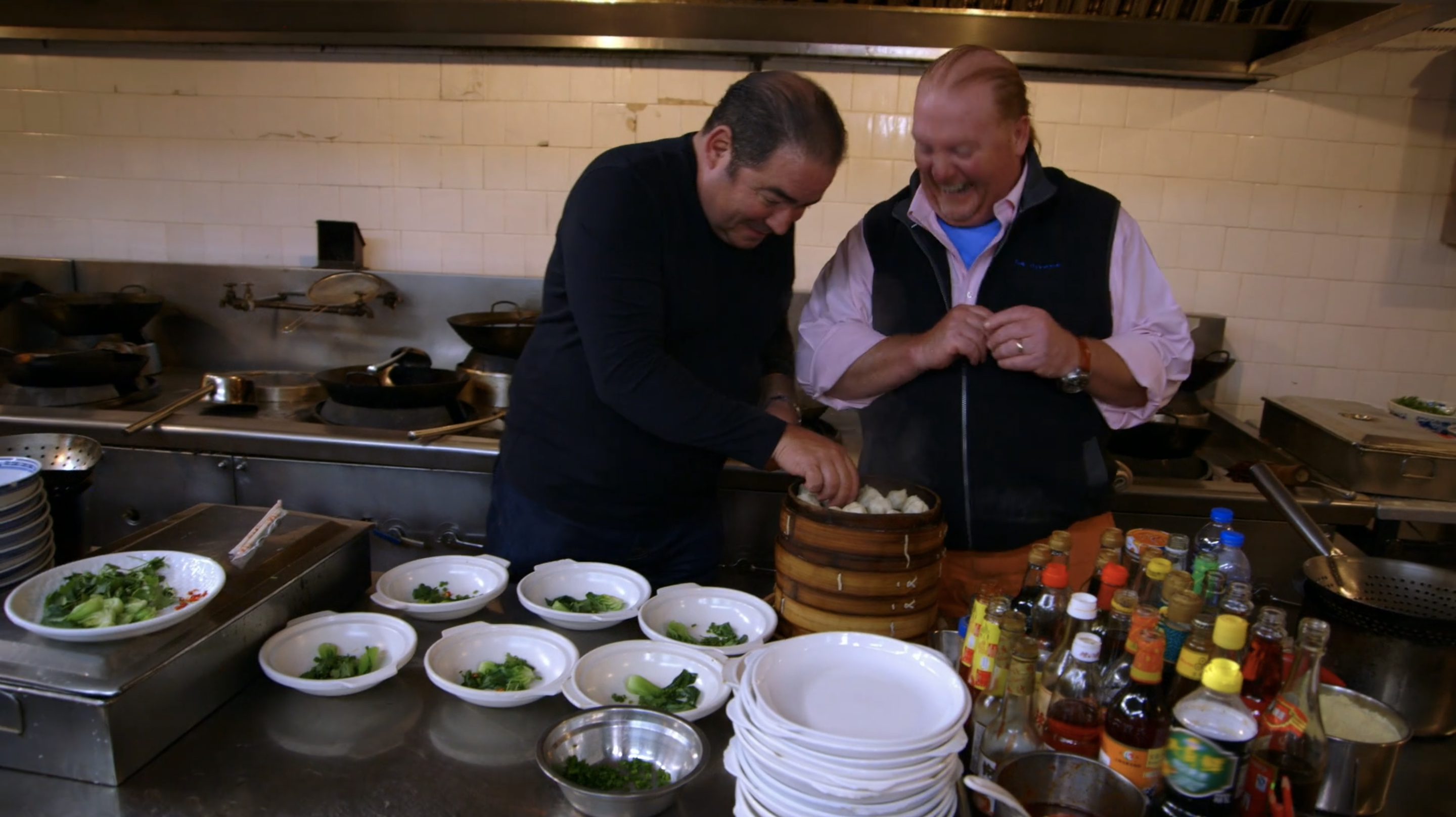}
    \caption{Background Image}
    \label{fig:light-rendering-background-image}
\end{subfigure}
\hfill
\begin{subfigure}{0.4\textwidth}
    \includegraphics[width=\textwidth]{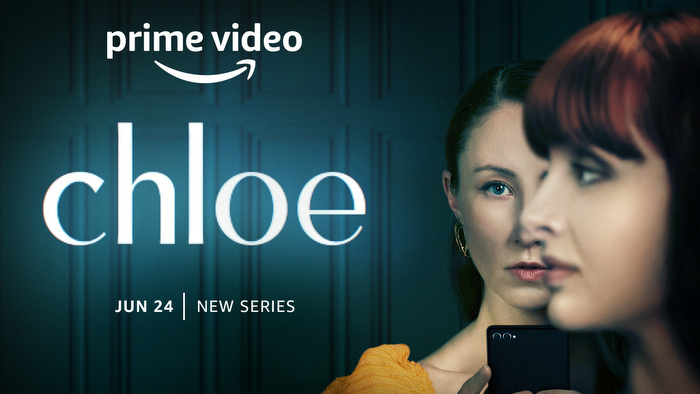}
    \caption{Ad}
    \label{fig:light-rendering-ad}
\end{subfigure}
\hfill
\begin{subfigure}{0.4\textwidth}
    \includegraphics[width=\textwidth]{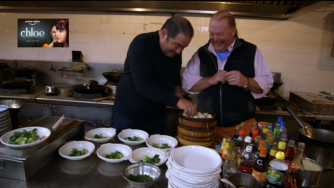}
    \caption{Color Transfer}
    \label{fig:light-rendering-color-transfer}
\end{subfigure}
\hfill
\begin{subfigure}{0.4\textwidth}
    \includegraphics[width=\textwidth]{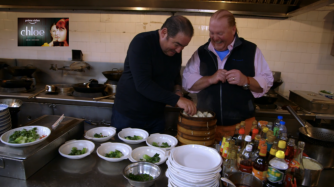}
    \caption{Histogram Matching}
    \label{fig:light-rendering-histogram-matching}
\end{subfigure}
\hfill
\begin{subfigure}{0.4\textwidth}
    \includegraphics[width=\textwidth]{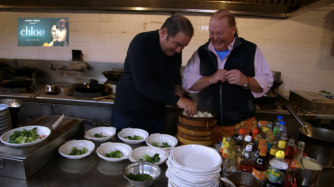}
    \caption{Brightness matching}
    \label{fig:light-rendering-brightness-matching}
\end{subfigure}
\hfill
\begin{subfigure}{0.4\textwidth}
    \includegraphics[width=\textwidth]{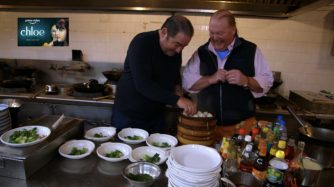}
    \caption{Light Transfer}
    \label{fig:light-rendering-light-transfer}
\end{subfigure}

\caption{Qualitative evaluation of ambient light rendering strategies.}
\label{fig:ambient-light-rendering}
\end{figure}

With lack of ground truth data and open source implements, we perform qualitative evaluation of the methods discussed in~\ref{subsec:ambient-light-rendering}. The LAB light transfer algorithm was chosen to be closer realistic illumination condition. We also check the cdf of pixel intensities of the ad image, video frame as well the render ad as show in figure~\ref{fig:ambient-light-rendering}.

\subsection{Ad placement}\label{subsec:ad-placement}
Due to the unavailability of labelled data with rendered image, we were unable to test the quantitative metrics of this task. The quality of the Ad image reduce while warping and rendering using OpenCV as it uses interpolation techniques. The rendering quality of Ad is better in high resolution image compared to low resolution image. For example, in the figure~\ref{fig:light-rendering-background-image} (cropped from original video frame of dimension 288X512) , the empty space location identified has dimension of 50x100 whereas the original Ad image dimension was 300X600. This brings about resizing to 6X smaller size. For larger resolution in~\ref{fig:light-rendering-background-image} (original dimension of 1080X1920), the empty space location identified has dimension of ~150x300 (2x smaller than original Ad image).


\subsection{Ad tracking}
The metric used was reprojection error which measures how far off in pixel coordinates, the Ad location is on previous image $t-1^{th}$ with regards to to its ground truth if we reverse the learnt transformation from current image $t^{th}$ location. We used the predictions from empty space location model as ground truth data. The metrics for top-2 feature matching algorithms (selected based on the \#matches generated) are displayed in table ~\ref{tab:reprojection-error-benchmark}. The lower the metric, better the pipeline is. There is no trend (feature detection/description) that deep learning models outperform classical techniques. While SuperPoint had the lowest error, Kornia had higher error than SIFT (classical).
%

\begin{table}[ht]
  \caption{Reprojection Error benchmark}
  \label{tab:reprojection-error-benchmark}
    \centering
    \small
    \begin{tabular}{llll}
        \toprule
        \textbf{Detection} & \textbf{Matching} & \textbf{Outlier filter} & \textbf{Reprojection error} \\
        \midrule
        Kornia LOTR & - & ransac & 0.798 \\
                    &  & magsac & 0.786 \\
        \midrule
        Superpoint (Pytorch) & match\_sym\_fginn\_intersection & ransac & 0.755 \\
         & match\_sym\_fginn\_intersection & magsac & 0.759 \\
         \midrule
        Superpoint (Pytorch) & match\_sym\_fginn\_union & ransac & 0.748 \\
         & match\_sym\_fginn\_union & magsac & 0.753 \\
         \midrule
        SIFT (OpenCV) & match\_sym\_fginn\_intersection & ransac & 0.763 \\
         & match\_sym\_fginn\_intersection & magsac & 0.818 \\
         \midrule
        SIFT (OpenCV) & match\_sym\_fginn\_union & ransac & 0.803 \\
         & match\_sym\_fginn\_union & magsac & 0.795 \\
         \midrule
        Orb (OpenCV) & match\_sym\_fginn\_intersection & ransac & 0.754 \\
         & match\_sym\_fginn\_intersection & magsac & 0.793 \\
         \midrule
        Orb (OpenCV) & match\_sym\_fginn\_union & ransac & 0.756 \\
         & match\_sym\_fginn\_union & magsac & 0.816 \\
         \bottomrule
    \end{tabular}
\end{table}

\subsection{ML Pipeline}\label{subsec:ml-pipeline}
Our VPP pipeline is an automated python script that call multiple models hosted on 4 different GPUs tested on an Amazon EC2 p2.8xlarge instance. This pipeline currently has an 5-6 FPS (frames per second) for low resolution videos (288X512) and 1-2 FPS for high resolution (1080X1920) videos.

\section{Limitations and Future Work}\label{sec:limitations-future-work}
We have identified the following areas for future exploration.

\subsection{Identifying suitable placement location}\label{subsec:identifying-suitable-placement-location3}
For an accurate empty space detection model, we recommend an exhaustive data annotation strategy which covers all possible empty spaces in a scene rather than a few. Additionally, we would recommend training the model over multiple image resolutions and over a larger annotated dataset for perspective aligned predictions.

\subsection{Kitchen-scene detection}\label{subsec:kitchen-scene-detection}
The current rule-based method is not 100\% accurate. In a False Negative scenario, the object won’t be rendered and may cause the ad to flicker. The models are highly confident ($\ge 90\%$) when at least upper half of the human body is visible. In edge cases where the camera covers other parts of the body the model might predict a False negative. Thus, the ad won’t be rendered even when the wall is empty. Collecting labelled dataset with different parts of human-body visible and indoor artifacts to train models with high accuracy can help in accurate classification of scene semantics.

\subsection{Occlusion Handling}\label{subsec:occlusion-handling2}
Virtual object will flicker if the image segmentation is not consistent. Most human segmentation models cannot capture details like hairline, nails, hats etc. Moreover the model’s performance degrades as the resolution of image decrease. Human segmentation results are not consistent when the person is partially present in the scene. We recommend expanding the occlusion detection to other kitchen objects like pan, bowl, spatula etc and exploring image matting techniques.

\subsection{Ambient Light Rendering}\label{subsec:ambient-light-rendering2}
Since there are no publicly available datasets or ML models for benchmarking ad rendering, we recommend creating a curated dataset of labeled dataset (background, ad, combined) which contains positive (lighting adjustment is good) and negative (lighting adjust is bad) samples, to start with. We also recommend experimenting with GAN architecture to create more realistic ads.

\subsection{Ad placement}\label{subsec:ad-placement3}
OpenCV based methods use interpolation techniques to warp/past image on to a location. This leads to small loss in resolution. This effect is highly evident in low resolution images compared high resolution ones. We don’t have any quantitative benchmarks on the extent of difference between CV based rendering vs high-definition rendering using softwares like Blend or Maya. Realistic rendering also involves placing the Ad on correct scale/dimensions that are consistent with 3D surroundings. This requires the knowledge of camera depth and scale of a known object which weren’t available to authors. We recommend testing and evaluating VFX applications for to compare ad rendering quality, and exploring single view-based camera calibration and depth estimation models for 3D scene understanding.

\subsection{Ad tracking}
Our homography estimation-based tracking is an approximation for small camera movements. Sudden camera movements will lead to distortion in rendering. Effectiveness of tracking is also based on the number of feature matches between 2 consecutive images. If the background is simple/plain or has highly reflective surface, the current pipeline will not able to distinguish different parts of image and can lead to poor matching. Homography based tracking can be used for static camera setting or in settings where the location of object is fixed and has visible markers (like 4 corners of goal post in a football game). Tracking based on 3-D world to 2-D understanding using camera calibration will have better accuracy than homography based estimation. This will not require the use of multi-step pipeline like that of homography based tracking. Benchmarking the effective of tracking and realistic rendering by learning camera parameters using multi-view camera or single view structure-from-motion algorithms on offline videos, can help understand the best strategy for tracking.

\section{Conclusion}\label{sec:conclusion}
In this paper, we present a solution for digitally placing a branded object into the scene of a movie or TV show. With our approach, advertisers can reach consumers without interrupting the viewing experience with a commercial break, as the products are seen in the background or as props. Our solution is easy to implement, requires minimal labeling, curation, supervision, and can be customized for various videos and advertisments. We hope the research community continue our work and develop better solutions for virtual product placement.

\vskip 0.2in
\bibliography{vpp}

\end{document}